\documentclass[conference]{IEEEtran}
\IEEEoverridecommandlockouts
\usepackage{cite}
\usepackage{amsmath,amssymb,amsfonts}
\usepackage{algorithmic}
\usepackage{graphicx}
\usepackage{textcomp}
\usepackage{hyperref}
\usepackage{xcolor}
\usepackage{booktabs}
\usepackage{subfigure}
\newcommand\setrow[1]{\gdef\rowmac{#1}#1\ignorespaces}
\newcommand\clearrow{\global\let\rowmac\relax}
\clearrow
\def\BibTeX{{\rm B\kern-.05em{\sc i\kern-.025em b}\kern-.08em
    T\kern-.1667em\lower.7ex\hbox{E}\kern-.125emX}}
\begin{document}

\title{Is it Fake? News Disinformation Detection on South African News Websites}

\author{\IEEEauthorblockN{Harm de Wet}
\IEEEauthorblockA{\textit{Department of Computer Science} \\
\textit{University of Pretoria}\\
Pretoria, South Africa\\
harmdewet01@gmail.com}
\and
\IEEEauthorblockN{Vukosi Marivate}
\IEEEauthorblockA{\textit{Department of Computer Science} \\
\textit{University of Pretoria}\\
Pretoria, South Africa \\
vukosi.marivate@cs.up.ac.za}
}

\maketitle

\begin{abstract}
Disinformation through fake news is an ongoing problem in our society and has become easily spread through social media. The most cost- and time-effective way to filter these large amounts of data is to use a combination of human and technical interventions to identify it. From a technical perspective, Natural Language Processing (NLP) is widely used in detecting fake news. Social media companies use NLP techniques to identify the fake news and warn their users, but fake news may still slip through undetected. It is especially a problem in more localised contexts (outside the United States of America). How do we adjust fake news detection systems to work better for local contexts such as in South Africa. In this work we investigate fake news detection on South African websites. We curate a dataset of South African fake news and then train detection models. We contrast this with using widely available fake news datasets (from mostly USA website). We also explore making the datasets more diverse by combining them and observe the differences in behaviour in writing between nations’ fake news using interpretable machine learning.
\end{abstract}


\section{Introduction}
Modern technology makes it easy for anyone with access to a cell phone or computer to publish anything online, regardless of expertise on the topics. For humans, it is getting harder to distinguish between factual news and fake news \cite{silverman.2016}. It is not feasible for online companies to have a team of people read everything that is being portrayed as news for validity. Although research has been done and tools, e.g. machine learning, are in place at some social media companies to identify and warn users \cite{mosseri2017}, there is still a lot of fake news on these platforms. This could be due to localisation or national differences in the way people write and/or conceal fake news. For example, the way South Africans and Americans speak and write English is not the same. South Africa further may have instances of code-switching \cite{mesthrie2013slang}. 

In this work, we investigate developing a machine learning model to detect South African fake news web-sites/content. We investigate the impact of using US fake news datasets against using a new dataset that is focused on South African fake news. For people to trust such a model to analyse news for them, the model should not be a black box where they cannot see how it classifies news as fake or factual. Thus, interpretable machine learning is used, highlighting on the articles’ text used for evaluation and testing how much which parts contributed to the prediction \cite{doshi2017towards}. The stylistic differences or behavioural characteristics between writing fake news and real news are also investigated in the interpretable articles when the model predicts on an article although it is not very obvious to spot these differences. We further release, with this paper, the South African fake news website dataset.

\section{Background}

A survey from 2016 by BuzzFeed \cite{silverman.2016} found that 75\% of American adults are fooled by the fake news headlines. If this study should reflect a wider part of the world’s population to a reasonable extent, the damage that fake news can cause becomes serious. It should be clear that fake news has the potential to ruin people’s lives and businesses. It should also be clear that fake news can be used for other agendas such as misleading people to vote for another party amongst other things. This is well documented as Information Disorder \cite{wardle2017information}. South Africa is not immune and has had its own challenges with disinformation \cite{wasserman2020fake}. Because of increased demand in fact-checking for fake news detection, multiple researchers have tried multiple Natural Language Processing (NLP) techniques to automate the task because of the large amount of data available. Fact-checking by a person requires researching and identifying evidence, understanding the information and what could be inferred from it which is a tedious task for everything that has to be checked. Thus a human has to be in the loop. 

Natural language processing is a range of computerised computational techniques used to analyse text linguistically such that a computer can achieve a human-like language processing for real world application. The text used to train, evaluate and use the model has to be in a language used by humans to communicate and not constructed for analytical purposes \cite{liddy2001natural}. There are multiple approaches to NLP, the two popular ones are statistical (mathematical) and machine learning (artificial intelligence) based approaches. Statistical approaches use mathematical techniques to model linguistic phenomena and only use observable data as a source of evidence. As part of machine learning we have neural network based approaches that take large amounts of text data into consideration when creating a model, taking into account the purpose of the model resulting in a more robust model \cite{liddy2001natural}.

Supervised learning, an artificial intelligence learning approach, learns a function that maps input (n-dimensional vectors) to their respective desired output \cite{russell2013artificial}. Typically, the training data is given to the model as example input-output pairs without any specific rules. To create the mapping function, neurons are created with some arbitrary values called weights. These weights are changed in the training process to ensure each example input maps to its desired output, or as close to it as possible. In most fake news detection models, supervised learning is the training method of choice. Using existing articles and or statements that has been classified by means of fact-checking \cite{vlachos-riedel-2014-fact}. A typical text classification problem, even more so for fake news detection, is that the evaluation of a claim requires more context than just the fact being evaluated. Another problem is that well-constructed sentences may also convey false information \cite{vlachos-riedel-2014-fact}, thus the model would need to be trained with more context in mind.

Another approach that has been tested is using information surrounding the article like the author, origin, and credit history of the author \cite{long2017fake}. The problem with this approach is that it can introduce bias to the model, thus we will only be using the text content of the articles. When the model does not have knowledge of the author, the model would not be able to successfully classify the article. Most research on fake news detection has been done using very specific attributes and only focusing on certain aspects of the problem. Various combinations of the following were used:

\begin{itemize}
\item Short political facts \cite{perezrosas2017automatic}, \cite{wang2017liar}.
\item Social media statements \cite{perezrosas2017automatic}, \cite{kai2017fake}.
\item Authors, location, affiliation, speaker title and credit history \cite{wang2017liar}, \cite{long2017fake}.
\item Number of followers, likes and comments (media platform dependent) \cite{kai2017fake}.
\item Authenticity \cite{conroy2015automatic}, \cite{singh2017automated}.
\item Tone, emotion \cite{conroy2015automatic}, \cite{singh2017automated}.
\item Lexical, syntactic, semantic and readability properties \cite{long2017fake}, \cite{perezrosas2017automatic}.
\end{itemize}

Interpret means to “present in understandable terms” \cite{merriaminterpret}. Which is an important aspect of machine learning as some countries require algorithms that make decisions on interpretations of humans to provide an explanation \cite{europeanjournal}. Interpretable machine learning can be divided into two broad categories, namely intrinsic interpretability and posthoc interpretability. Intrinsic interpretability \cite{du2019techniques} is the art of constructing the model itself to explain what it does. For intrinsic interpretability there are some trade-offs that one needs to be aware of, between prediction accuracy and interpretability. Post-hoc interpretability \cite{du2019techniques} is the art of creating a second model that can explain the main model by modelling what the main model has done then attempt to explain what the main model has done. This method only estimates what the main model had done but there are no trade-offs in the main model’s performance or accuracy.

We investigate the stylistic differences in writing between nations with machine learning in areas where it could play an important role, such as fake news detection.

\section{Dataset Creation and Exploration}

\subsection{South African Dataset}

\begin{table}[t]
    \caption{Dataset Summary Statistics}
    \label{tab:summary_statistics}
\resizebox{\columnwidth}{!}{%
    \centering
    \begin{tabular}{lllll}
    \toprule
    & \textbf{SA Dis (SA 1)} & \textbf{SA News (SA 1)} & \textbf{US Dis (US 1)} & \textbf{US News (US 1)}\\
 \midrule
Article Count & 807 & 1614 & 3164 & 3171 \\
Mean Article Length & 375.66 & 460.36 & 679.13 & 873.26\\
Min Article Length & 0 & 36 & 0 & 7\\
Max Article Length & 2388 & 5687 & 20891 & 7602\\
Articles Above Mean & 272 & 558 & 995 & 1300\\
Articles Above 500 & 193 & 487 & 1344 & 2262\\
Unique Tokens & 341801 & 849737 & 2480838 & 3227398\\
Unique Tokens Per Article Mean & 191.50 & 231.13 & 313.45 & 397.99\\
Unique Tokens Per Article Min & 0 & 27 & 0 & 7\\
Unique Tokens Per Article Max & 885 & 1227 & 3942 & 2487\\
    \bottomrule
    \end{tabular}
}

\end{table}

We used, as sources, investigations by the news websites MyBroadband \cite{mbb2017} and News24\footnote{\url{https://exposed.news24.com/the-website-blacklist/}}. These articles covered investigations into disinformation websites in South Africa in 2018. They compiled lists of websites that were suspected to be disinformation. During the period from those articles to present, a number of the websites have become inaccessible or offline. We attempted to use the internet archives WayBack Machine\footnote{\url{https://archive.org/web/}} we could only get partial snapshots and error messages.

A web-scraper only worked for one of the sources although manual editing was still required to clean the text from Javascript code and some paragraph duplicates. On most of the other websites, a web-scraper did not work well as there were too many advertisements and broken parts of pages. Because of all these problems, most of the articles were manually copied and pasted and cleaned in flat files. In some cases, the text of articles could not be copied and was not made part of the South African disinformation corpus.

\begin{figure}
    \centering
    \subfigure[Top 50 words in SA Information]
    {
        \includegraphics[width=\columnwidth,trim = 40 20 40 20]{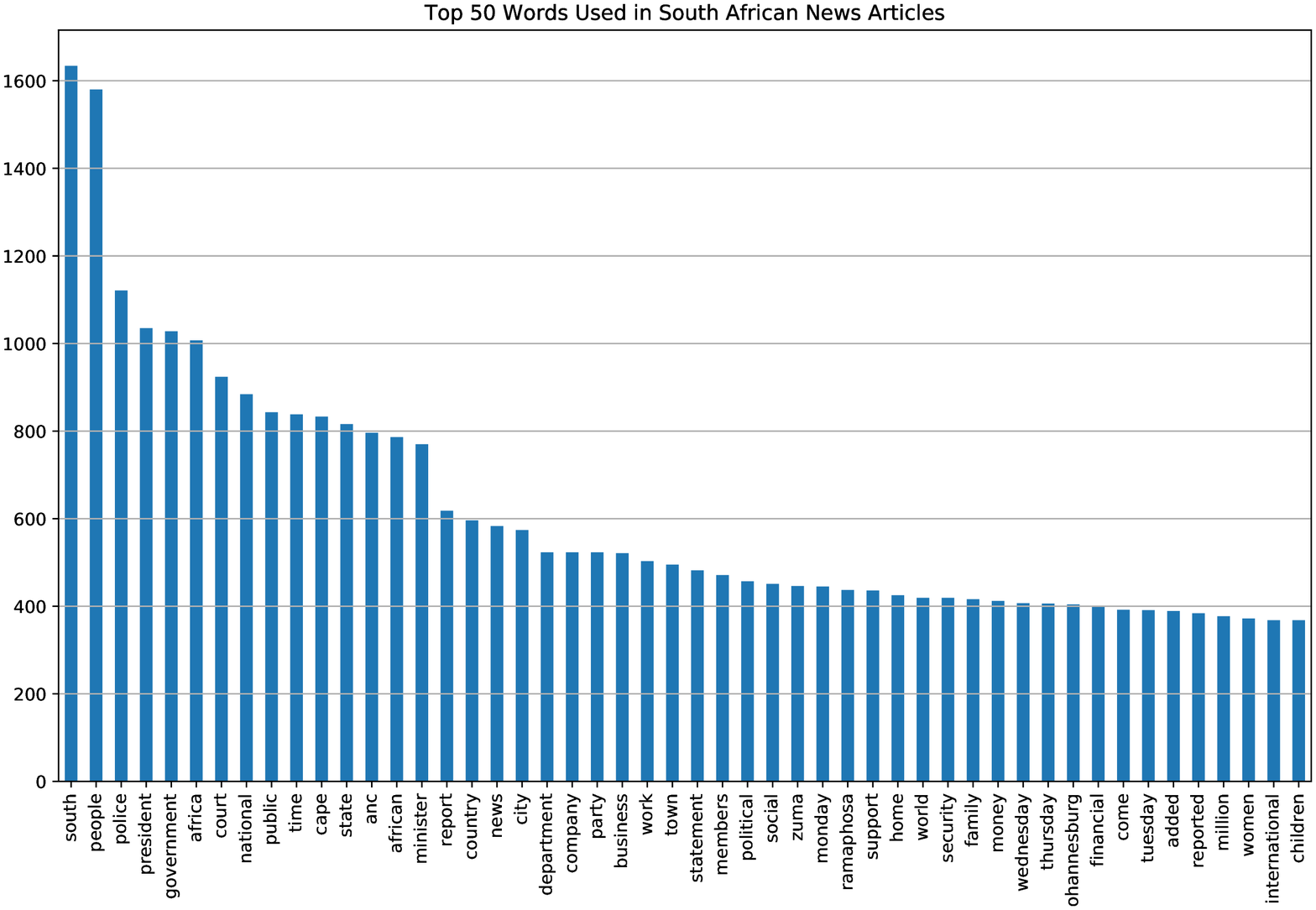}
        \label{fig:sa_informationtop}
    }
    \\
    \subfigure[Top 50 words in SA Disinformation]
    {
        \includegraphics[width=\columnwidth,trim = 40 20 40 20]{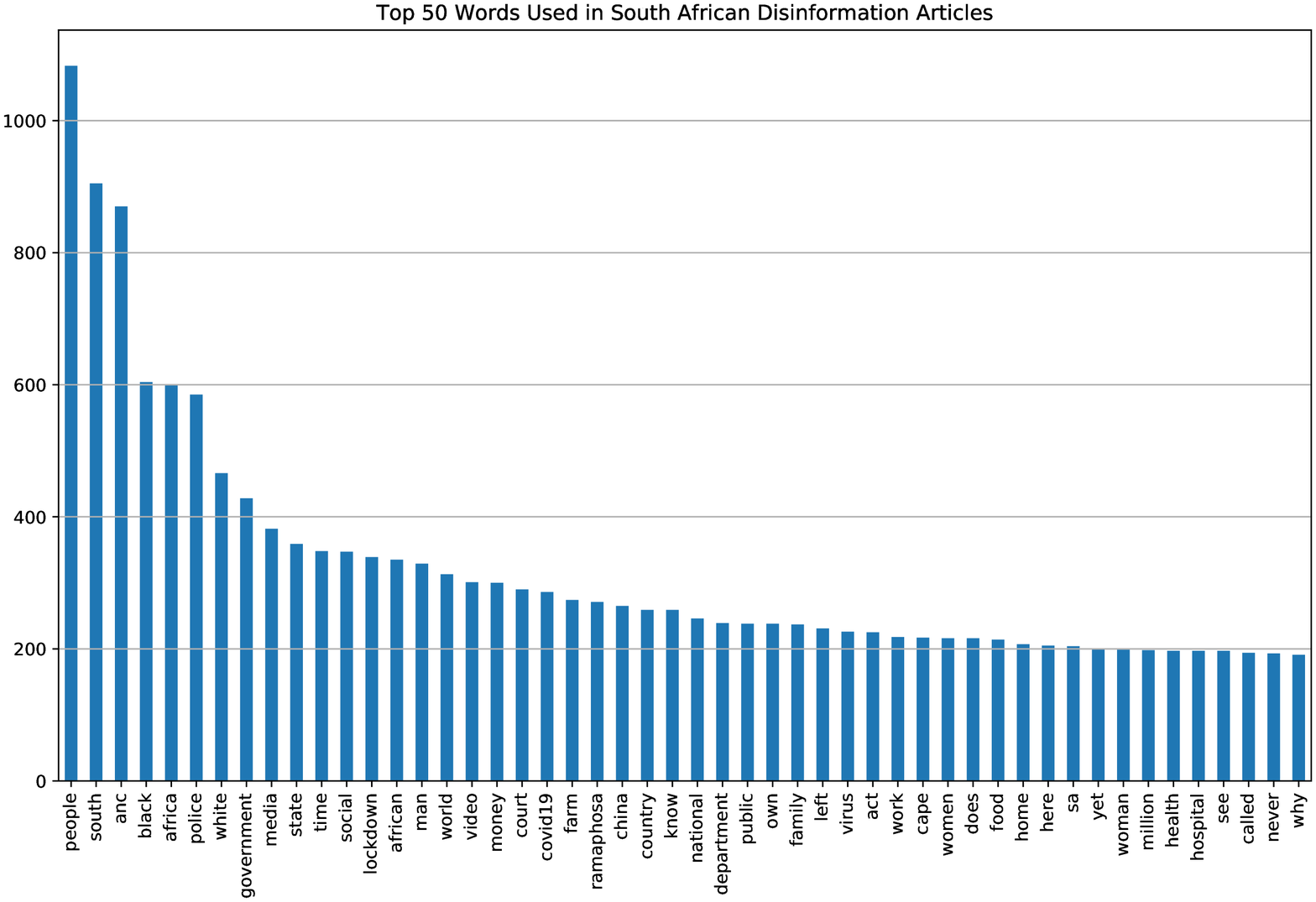}
        \label{fig:sa_disinformationtop}
    }
    \caption{Word frequency summaries for South African Data Set (SA1)}
    \label{fig:sa_summary}
\end{figure}

Real or truthful South African news used for training and evaluation of models was sourced from Media Monitoring Africa\footnote{\url{https://mediamonitoringafrica.org/}}, an independent organisation that monitors South African news agencies. These articles were already cleaned, noise removed and duplicates extracted. The real and fake news from South Africa was combined to form the SA 1 dataset. You can see the summary statistics for the SA corpus data in Fig \ref{fig:sa_summary}. We make available the South African fake news website data available through an online repository \url{https://github.com/dsfsi/sa-fake-news-2020} or \url{https://doi.org/10.5281/zenodo.4682843} for others to be able to use.

\begin{figure}
    \centering
    \subfigure[Disinformation SA Dataset Wordcloud]
    {
        \includegraphics[width=0.45\columnwidth]{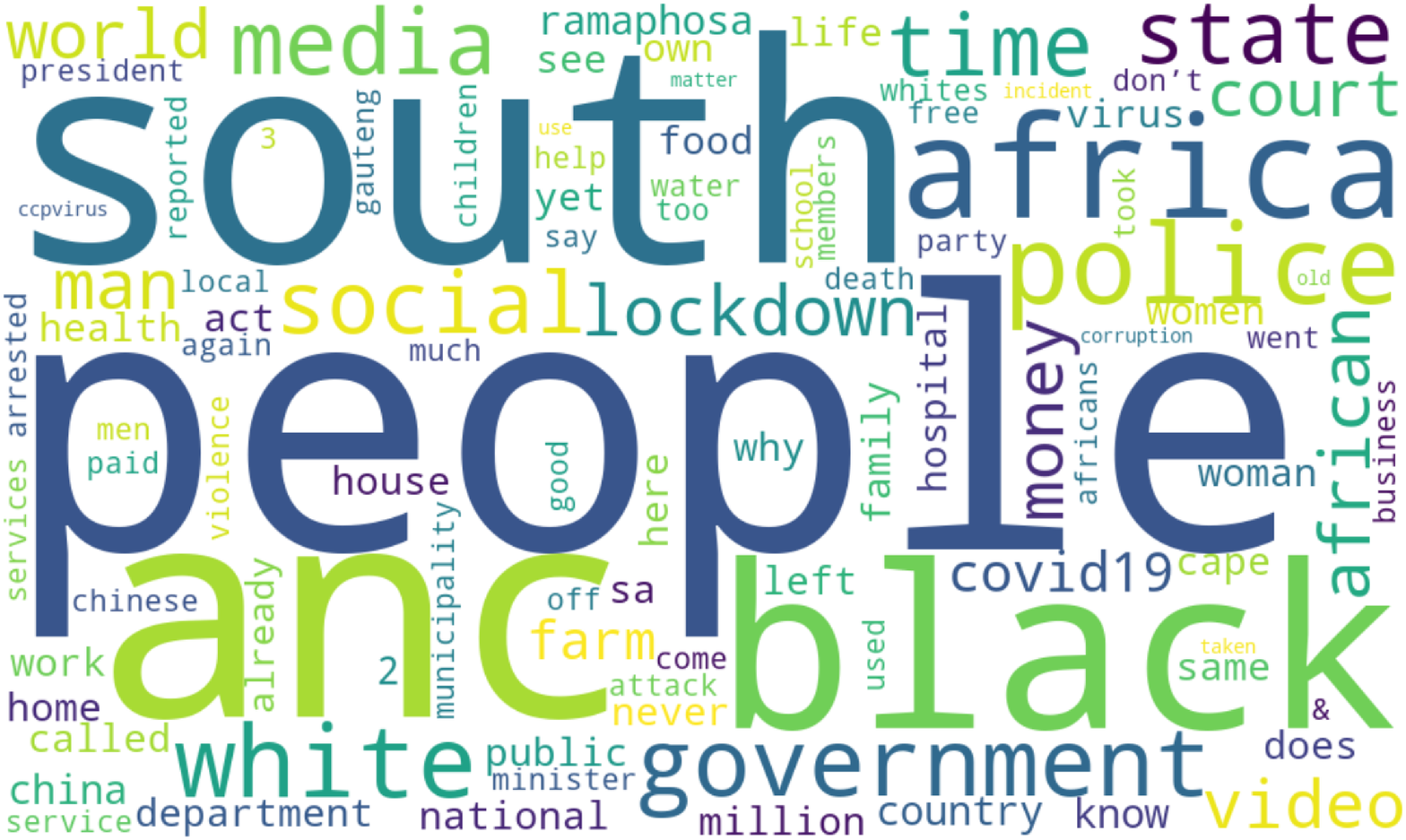}
        \label{fig:sa_disinformation}
    }
    \subfigure[Disinformation USA Dataset Wordcloud]
    {
        \includegraphics[width=0.45\columnwidth]{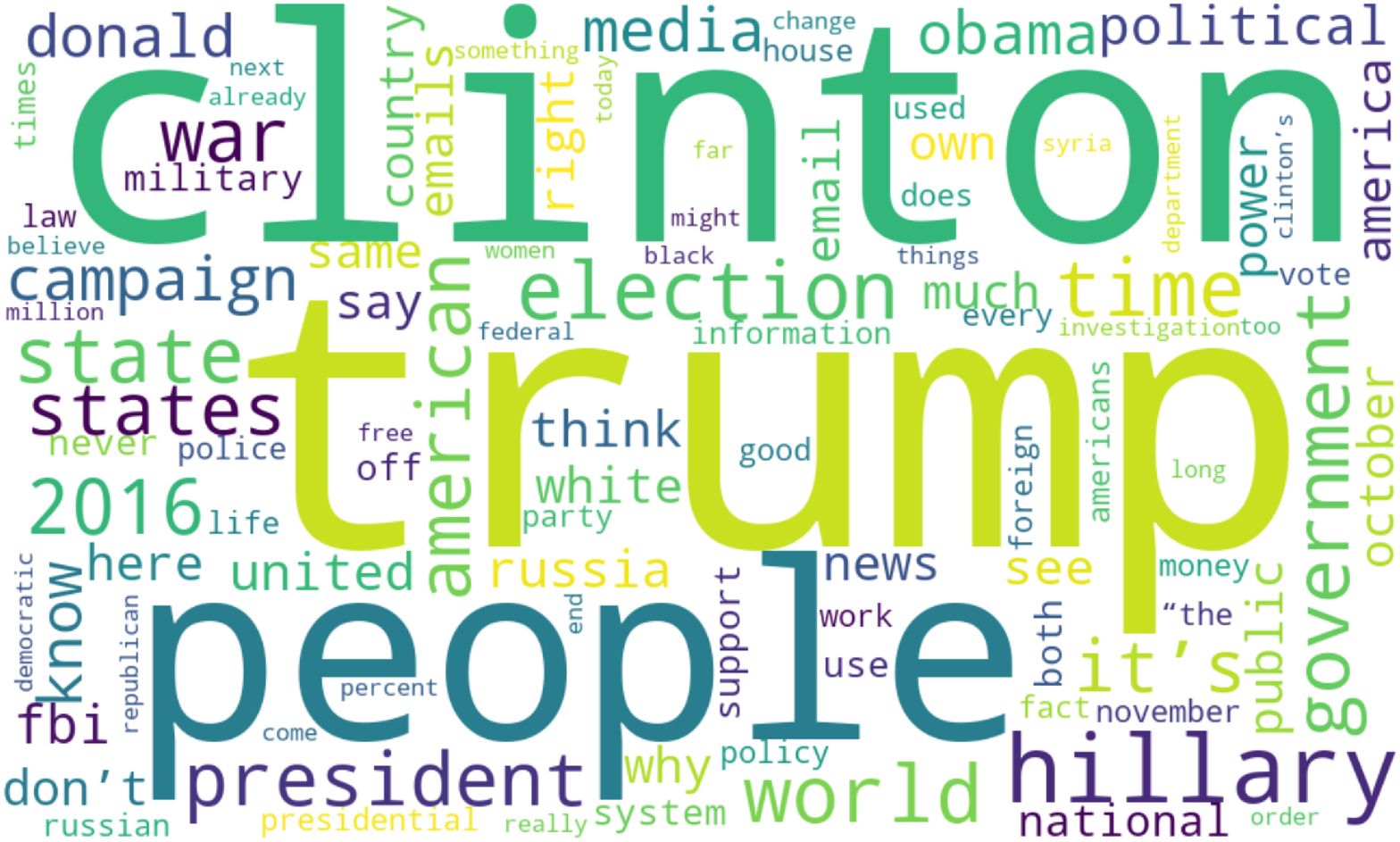}
        \label{fig:usa_disinformation}
    }
    \caption{Disinformation Wordclouds}
    \label{fig:wordclouds}
\end{figure}

% \begin{figure}
%     \centering
%     \includegraphics[width=\columnwidth,trim = 40 40 40 40]{figures/visualisations/eps/sa_disinformation_top50.eps}
%     \caption{Top 50 words in SA Information}
%     \label{fig:sa_informationtop}
% \end{figure}

% \begin{figure}
%     \centering
%     \includegraphics[width=\columnwidth,trim = 40 40 40 40]{figures/visualisations/eps/sa_disinformation_top50.eps}
%     \caption{Top 50 words in SA Disinformation}
%     \label{fig:sa_disinformationtop}
% \end{figure}

% \begin{figure}
%     \centering
%     \includegraphics[width=\columnwidth]{figures/visualisations/eps/sa_disinformation_cloud.eps}
%     \caption{Disinformation SA Dataset Wordcloud}
%     \label{fig:sa_disinformation}
% \end{figure}

\subsection{United States of America (USA) Dataset}

The US datasets were used to evaluate how our approach to the South African disinformation detection could generalise across different news and disinformation sources as well as the other way round. Further, the USA data was used in combination with the \emph{SA corpus} to create a combined model that straddles the two countries.

\begin{figure}
    \centering
    \subfigure[Top 50 words in US Information]
    {
        \includegraphics[width=\columnwidth,trim = 40 20 40 20]{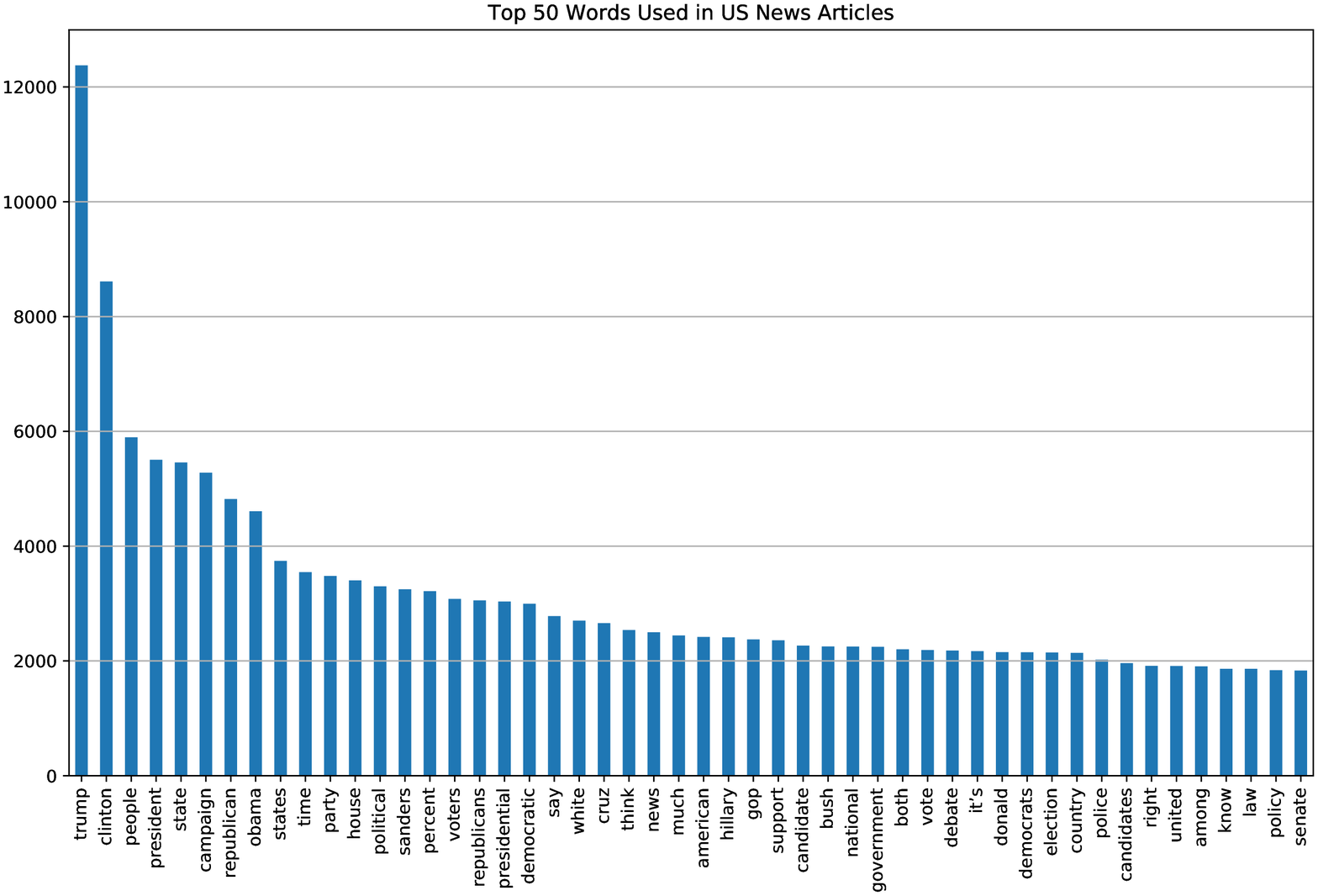}
        \label{fig:usa_informationtop}
    }
    \\
    \subfigure[Top 50 words in US Disinformation]
    {
        \includegraphics[width=\columnwidth,trim = 40 20 40 20]{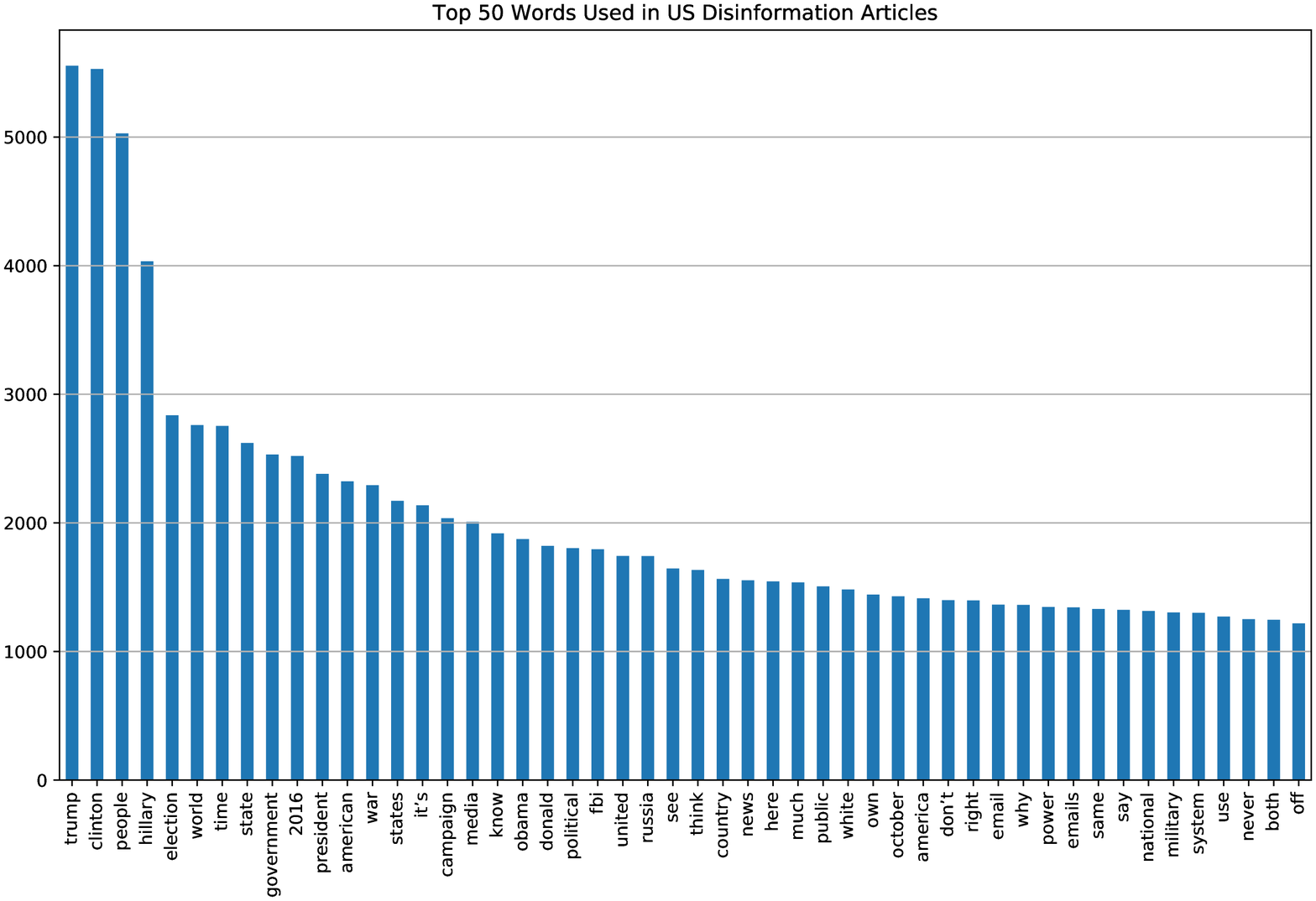}
        \label{fig:usa_disinformationtop}
    }
    \caption{Word frequency summaries for United States corpus (US)}
    \label{fig:usa_summary}
\end{figure}

We have two USA datasets, \emph{US 1} and \emph{US 2}. \emph{US 1} data contains a scaled down dataset that originated from a Kaggle Dataset on \emph{Getting Real about Fake News} \cite{kaggle2016}. The scaled down dataset \cite{odsc2017} contains equal amounts of real and fake news data and is available on GitHub\footnote{\url{https://github.com/lutzhamel/fake-news}}. The \emph{US 2} data is also from Kaggle, names \emph{Fake and real news dataset} \cite{kaggle2020}. The dataset has more data but upon inspection the data was found to contain more noise than \emph{US 1}. We use it for generalisation evaluations. You can see the summary statistics for the US corpus data in Fig \ref{fig:usa_summary}.

\section{Modelling}

There are three aspects to this research that build on top of each other: \begin{itemize}
    \item fake news classification,
    \item stylistic understanding,
    \item  and making the results interpretable such that a person can understand the model’s prediction.
\end{itemize}

To evaluate the stylistic understanding of the models, two models of each neural network type were trained, one with \emph{SA corpus} and the other with \emph{US corpus} then each model’s test data was used to evaluate the other model trained with the different nation’s data.

\subsection{Pre-processing}
The text is cleaned from all punctuation and capitalization as it is assumed that it does not contribute to the context and behaviour in the article. Stopwords have also been removed from the text for some of the models stopwords do not contribute enough to the overall performance of the models and that it reduced the dataset size and computation time with minimal, if any, accuracy loss \cite{saif2014stopwords,silva2003importance,schofield2017pulling}.

For the base models we used Term Frequency — Inverse Document Frequency (TF-IDF) encoding for the words, the length of the articles was not modified. For the Long short-term memory (LSTM) models, the first 500 words were used in training and testing. The mean length of each corpus is less than 500 words (see Table I). A tokenizer was used to encode the words, splitting the words into an array, turning each word into its numeric position in the word dictionary. For all articles under 500 words in length, the encoded list of vectors were padded with zeros.

Text augmentation was used in some cases to expand the limited dataset. From the accuracy table, it was shown that increasing the training data using augmentation proved to be effective \cite{marivate2020improving}. WordNet based augmentation was used with the \textit{TextAugment} library\footnote{\url{https://github.com/dsfsi/textaugment}}. 

\subsection{Models}

\subsubsection{Base Models}
The starting point of the neural network experiments was to get a simple baseline model to evaluate the output of the more complicated models. Logistic regression with a TF-IDF bag-of-word encoding. For Logistic Regression and Passive Aggressive Classifier base models, the Scikitlearn\footnote{\url{https://scikit-learn.org/}} libraries were used. These libraries determine the layers and nodes required for the models itself.

\subsubsection{LSTM Models}
For the LSTM models Keras was used and each layer customised, and layers put together to form the model.
Word2Vec \cite{mikolov2013distributed} was used to create a matrix of word representations for every word in the dictionary that is used by the tokenizer. These embeddings formed a matrix of values that became the weights for the untrainable embedding layer that receives the padded token sequences as input and turns it into the embedded representation of the word. The embedding layer made it possible to have a smaller memory footprint since it was not required to have the trained Word2Vec model in memory to embed words before the model can be used. The drawback of this approach was that words that are not in the dictionary but in the text that was predicted, will not be used as input for the prediction.

The embedding layer was followed by a dropout layer to add regularization to the model when training. Then, followed by the convolutional is the max pooling layers used to extract features and reduce the size of the model. Finally, the layers are followed by 2 LSTM layers ending in 3 layers feed-forward fully connected layers. The following activation functions were attempted in some of the layers of the LSTM models: Hyperbolic Tangent (TanH), Yann LeCun TanH, Elliot, Rectified Linear Unit (ReLU) and Sigmoid. Only ReLU and Sigmoid, in combination, produced report worthy results.

\section{Experiments}
The data used to train the Logistic Regression models consisted of balanced classes. Training with the American corpus produced a better accuracy score than the model trained with the South African corpus. Predicting classes for the \emph{US corpus} using the \emph{SA corpus} trained model produced poor results. Refer to Table \ref{tab:performance}. The F1 score for this experiment is 0.564 with a per class accuracy of 0.431 for fake news articles and 0.696 for true news articles. The drop in accuracy was expected but turned out to be more severe than the expectation.  Predicting classes for the \emph{SA corpus} using the \emph{US corpus} trained model also produced poor results but even more so for classifying the fake news itself. With an F1 score of 0.473 and a per class accuracy of 0.260 for the fake news articles and 0.686 for the true news articles. 

To preserve the context of word sequences, LSTM neural networks were constructed. To create an LSTM model baseline, equal numbers of articles of fake news and real news were used to train the models. The balanced classes trained and tested LSTM model trained with \emph{SA corpus} achieved 0,520 accuracy and the \emph{US corpus} trained model achieved 0,895 accuracy.

To increase the accuracy of the first LSTM models, a data imbalanced model was trained with the \emph{SA corpus}, since there are more real than fake news in the real world, reaching an accuracy of 0,755. No training data imbalance was introduced to the American trained model as the data source only contained half of each class.

To increase the accuracy of the \emph{SA corpus} trained LSTM model, Wordnet based augmentation \cite{marivate2020improving} was used to expand the dataset. The increased dataset size improved the model’s accuracy score from 0,755 to an average of 0,845. Augmenting the \emph{US corpus} also increased the American trained models accuracy from 0,895 to an average of 0,968.

Combining the \emph{SA corpus} and the \emph{US corpus (US 1)} for training and testing another model, the model reached an average accuracy score of 0.884 and a peak of 0.906 which is better than testing different nations’ data on different models. The results were harder to analyse for behavioural characteristics and the accuracy a bit lower, but it performed better than training models with one nations data and testing with another nation’s data. The final LSTM models were trained 5 times to get an average accuracy score. Refer to Table \ref{tab:performance} for results..

\begin{table}
\caption{Experiment results (bolded rows referred to in text)}
\label{tab:performance}
\resizebox{\columnwidth}{!}{%
    \centering
    \begin{tabular}{llllll}
    \toprule
    Model & Data Trained & Data Tested & Average Accuracy & F1 Score & Per Class Accuracy\\
 \midrule
Logistic Regression (Base) & US1 Subset1 & US1 Subset2 & 0,919 & 0,916 & Fake: 0,923 | Real: 0,914\\
\setrow{\bfseries}Logistic Regression (Base) & US1 Subset1 & SA1 & 0,559 & 0,473 & Fake: 0,260 | Real: 0,686\\
Logistic Regression (Base) & SA1 Subset1 & SA1 Subset2 & 0,824 & 0,824 & Fake: 0,831 | Real: 0,816\\
\setrow{\bfseries}Logistic Regression (Base) & SA1 Subset1 & US1 & 0,604 & 0,564 & Fake: 0,431 | Real: 0,696\\
\setrow{\bfseries}LSTM (Base - Balanced Classes) & SA1 Subset1 & SA1 Subset2 & 0,52 & 0,335 & Fake: 0,668 | Real: 0,000\\
\setrow{\bfseries}LSTM (Base - Balanced Classes) & US1 Subset1 & US1 Subset2 & 0,895 & 0,895 & Fake: 0,899 | Real: 0,891\\
\setrow{\bfseries}LSTM (Class Imbalance) & SA1 Subset1 & SA1 Subset2 & 0,755 & 0,757 & Fake: 0,651 | Real: 0,811\\
LSTM (Class Imbalance) & SA1 Subset1 & US1 & 0,492 & 0,492 & Fake: 0,493 | Real: 0,490\\
LSTM & US1 Subset1 & SA1 & 0,706 & 0,475 & Fake: 0,439 | Real: 0,549\\
\setrow{\bfseries}LSTM (Wordnet Augmentation) & SA1 Subset1 & SA1 Subset2 & 0,845 & 0,833 & Fake: 0,774 | Real: 0,862\\
LSTM (Wordnet Augmentation) & SA1 Subset1 & US1 & 0,61 & 0,61 & Fake: 0,610 | Real: 0,612\\
\setrow{\bfseries}LSTM (Wordnet Augmentation) & US1 Subset1 & US1 Subset2 & 0,968 & 0,968 & Fake: 0,968 | Real: 0,968\\
LSTM (Wordnet Augmentation) & US1 Subset1 & SA1 & 0,583 & 0,56 & Fake: 0,607 | Real: 0,556\\
\setrow{\bfseries}LSTM (Wordnet Augmentation) & SA1 and US1 Subset1 & SA1 and US1 Subset2 & 0,884 & 0,906 & Fake: 0,897 | Real: 0,913\\
LSTM & US2 Subset1 & US2 Subset2 & 0,961 & 0,961 & Fake: 0,961 | Real: 0,960\\
LSTM & US2 Subset1 & SA1 & 0,658 & 0,657 & Fake: 0,745 | Real: 0,481\\
    \bottomrule
    \end{tabular}
}
\end{table}

\section{Model Prediction Interpretability}

\begin{figure}
    \centering
    \includegraphics[width=\columnwidth]{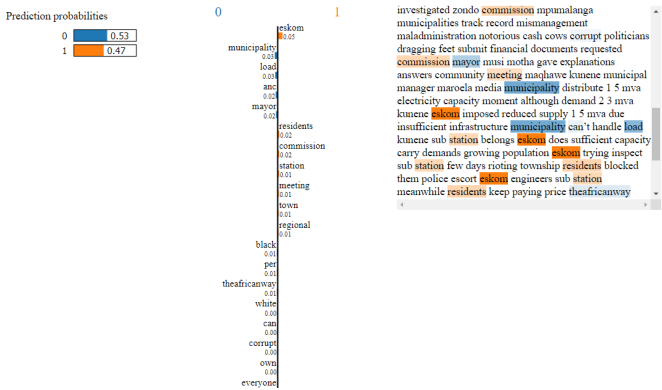}
    \caption{Lime explainations using LIME}
    \label{fig:lime}
\end{figure}

The logistic regression models’ results were made interpretable using Local interpretable model-agnostic explanations (LIME) \cite{ribeiro2016should}. Analysing the articles with the LIME explanations showed that the model relies on the presence and counts of specific keywords to make its prediction, not taking enough context into account. The word based reliability can be seen in the Fig \ref{fig:lime}.

\begin{figure}
    \centering
    \subfigure[Intrinsically interpreted weights used to highlight words and phrases according to the weights.]
    {
        \includegraphics[width=\columnwidth]{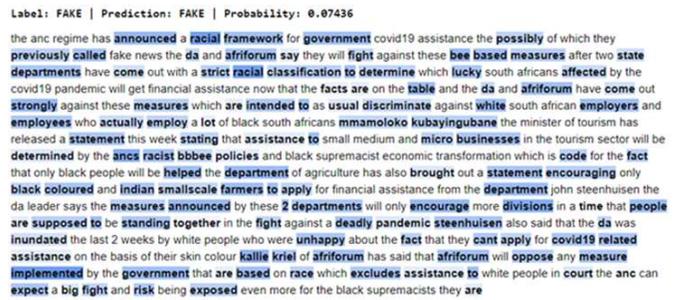}
        \label{fig:lstm1}
    }
    \\
    \subfigure[Top 20 words that contributed to the classification and their contributed weight.]
    {
        \includegraphics[width=\columnwidth]{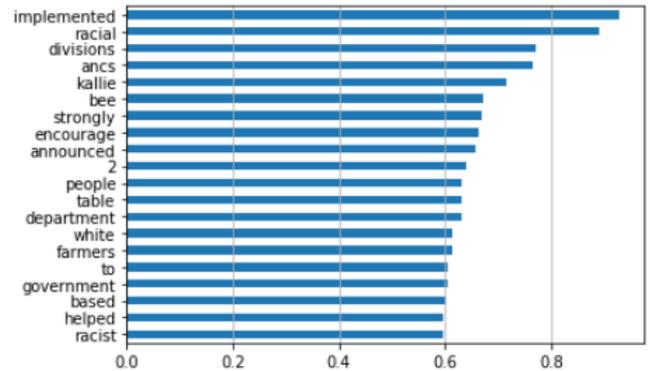}
        \label{fig:lstm2}
    }
    \caption{LSTM interpretability visualisations}
    \label{fig:lstm_interpretable}
\end{figure}

Intrinsic interpretability was used to make these LSTM models interpretable. Two permutation layers were added and the weights of the second permutation layer was extracted from the model. These weights were used as values to indicate how much which words in the text contributed to the classification. When a piece of text was processed by the model, a breakdown of how much each word contributed to the classification was given as a bar graph with only the contributing words and the text itself. The words are highlighted lighter or darker according to the weight that it contributed to the classification. See Fig \ref{fig:lstm_interpretable}. 

% \begin{figure}
%     \centering
%     \includegraphics[width=\columnwidth]{figures/Graphics/eps/intrinsic_text.eps}
%     \caption{Intrinsically interpreted weights used to highlight words and phrases according to the weights.}
%     \label{fig:lstm1}
% \end{figure}

\subsection{Discussion and observations}

% The main examples used to analyze the interpretable result of the model can be found in Appendix B.

The following were the behavioural factors contributing to the classification for the trained models

\subsubsection{\emph{SA corpus} models}

There was sentence repetition in some of the fake news that was not found in the real news. The fake news contained more exaggerated words and phrases than real news and in some articles multiple exaggerations for example:
\begin{itemize}
    \item “billion rand taken tenders worth billions” - article 3
    \item “deadly pandemic” - article 4
    \item “deep loss” - article 5
    \item “unbearable pain” - article 5
    \item “work tirelessly” - article 5
    \item “wanton violence” - article 5
    \item “deep pain” - article 5
\end{itemize}

In some cases it seems like the model takes simplified words and speech as an indication that an article is fake news, this could be the reason why some real articles have been classified as fake, for example:
\begin{itemize}
    \item “hes giving all thanks” - article 1
    \item “fix issues” - article 3
    \item “lucky” - article 4
    \item “employers and employees who actually employ” - article 4
    \item “crash”- article 1
\end{itemize}

Counter examples for the simplified words and speech:
\begin{itemize}
    \item “selection criteria focussed on” - article 7
    \item “own constitution on eligibility” - article 7
    \item “warned defaulting municipalities” - article 9
\end{itemize}

In some articles phrases highlighting the biased and opinionated behaviour of the writer was used as contributing factor to a fake news classification for example:
\begin{itemize}
 \item “da understands better than anyone” - article 3
 \item “people are supposed to” - article 4
\end{itemize}

\subsubsection{\emph{US corpus} models}
There is fewer repetition of sentences than in the South African fake news, but in most cases with repetition, the second and third time the same sentence is present, the wording is slightly changed. The fake news was also found to be more opinionated than the real news for example one  could see the author referred to themselves as being part of some group where injustice occurred.
“these people must think the Americans are stupid” - article 4

There was also some places where it seemed like the model used exaggeration as part of the classification of fake news:
“have seen countless aggravated assaults” and “down right murders” -article 4. But then there were also cases where a fake article was marked as real with high probability where exaggeration is present like in article 13. In both South African and American fake news, it appears that third person speech is more prevalent in fake news than in real news and first person speech more prevalent in real news than in fake news.

Also found in both nations’ data is racial words and expressions and curse words, with the correct and the wrong spelling, that the models used to distinguish the fake and real news from each other as a basis to classify articles as fake news, but not in all cases. One nation’s model performed poorly classifying another nations data because the data used for the training might not be diverse enough and the different nations’ datasets news categories also do not overlap enough. After an analysis of the two different \emph{US corpus}, the news diversity was poor since the model trained with the one set performed poorly on the other set, however in the same set, the models achieved good prediction accuracy on the test data because we have a lot of data for each source.

The South African model also reached acceptable results with the \emph{SA corpus}. This model could reach better performance with more data that is more diverse data. Better manual classification and inspection of training data could be part of the reason why the \emph{US corpus} trained models performed better on their own data but poor on the other nation’s data. The model where both nations data was used together for training and testing also presented similar interpretable output with the same behavioural characteristics although it was difficult to spot them as it seemed a bit more random.

\section{Conclusion}

In this work we have investigated the creation of fake news website detection models using a new South African dataset as well as two US datasets. We further provide insight into the behavioural factors that may lead the models to classify the websites as fake or not. This is just the beginning of providing models that are better suited for South Africa. We also release the South African fake news dataset as an online dataset for others to use and also expand on. There is still more work to be done. Investigations into changes into how fake news spread in South Africa is still to be done. What happens when the websites themselves change how they write to avoid detection. The archiving of past fake news websites is important as they tend to disappear after a few years. As such there is a need for archiving tools for such websites as they provide good data for future models.

\section{Acknowledgements}

The authors would like to thank Media Monitoring Africa for their support. Further, we would like to thank the reviewers for their feedback and Tshephisho Sefara for proof reading. 

\bibliographystyle{IEEEtran}
\bibliography{references_africonfake}

\end{document}